\documentclass[11pt,a4paper]{article}
\usepackage[hyperref]{emnlp-ijcnlp-2019}
\usepackage{times}
\usepackage{latexsym}
\usepackage{bbm}
\usepackage{amsmath}
\usepackage{multicol}
\usepackage{graphicx}
\usepackage{makecell}
\graphicspath{{./images/}}

\usepackage{url}

\aclfinalcopy 

\title{Explaining Sequence-Level Knowledge Distillation\\as Data-Augmentation for
Neural Machine Translation}

\author{Mitchell A. Gordon \\
  Johns Hopkins University \\
  {\tt mitchg@jhu.edu} \\\And
  Kevin Duh \\
  Johns Hopkins University \\
  {\tt kevinduh@cs.jhu.edu} \\}

\date{}

\begin{document}
\maketitle
\begin{abstract}
  
  Sequence-level knowledge distillation (SLKD) is a model compression technique that leverages large, accurate teacher models to train smaller, under-parameterized student models.
  Why does pre-processing MT data with SLKD help us train smaller
  models? We test the common hypothesis that SLKD addresses a capacity
  deficiency in students by ``simplifying'' noisy data points and find it unlikely in our case.
  Models trained on concatenations of original and  ``simplified''
  datasets generalize just as well as baseline SLKD. 
We then propose an alternative hypothesis under the lens of data augmentation and regularization.
  We try various augmentation strategies and observe that dropout regularization can become unnecessary. Our methods achieve BLEU gains of 0.7-1.2 on TED Talks.  
\end{abstract}

\section{Introduction}
The recipe for success in modern machine learning involves training big neural
networks on big data. This is especially true in Neural Machine Translation (NMT), where models regularly
take weeks to train and cost gigabytes of disk space. Attempting to reduce this bloat
has led to much research on model compression \cite{model-compression}, which takes large
networks and makes them smaller. Small models can save money on compute time,
deploy on mobile devices, and, optimistically, give us insight into the nature
of our learning tasks and the limitations of our optimization algorithms.

We need model compression, in general, because small neural networks tend to be
harder to train than big ones. In an ideal world, our models would size
themselves \cite{kenton}, and our optimization algorithms would find the best
weights possible, no matter what hyperparameters we pick. However, explanations
for why stochastic gradient descent (SGD) on big neural nets works so well remain elusive, although
significant steps are being taken \cite{vidal,gunesekar}.

Here, we investigate a compression technique called sequence-level knowledge
distillation (SLKD) and attempt to discern why it improves the performance of small
models. We hypothesize that, contrary to common belief, SLKD does not always address a
capacity deficiency in models by removing problematic data points. Instead, we
think it can act like regularization, guiding SGD towards generalizable
solutions by placing more data points on the teacher's learned manifold.

\section{Background}
\textbf{Machine Translation}
Successful machine learning starts with a good loss function, which is both art
and science. The loss function determines what our ideal world looks like and
ultimately what our models learn.

For example, consider the usual machine translation task: we have some pairs of
source and target sentences $(\mathbf{t_i},\mathbf{s_i})$ from a data
distribution $\mathcal{D}(\mathbf{t},\mathbf{s})$ such that
$(\mathbf{t_i},\mathbf{s_i})\sim\mathcal{D}(\mathbf{t},\mathbf{s})$. Sentences
are sequences of words $\mathbf{s_i} = [s_1, ..., s_I]$, $\mathbf{t_i} = [t_1,
..., t_J]$ with vocabulary $\nu$. We'll denote the space of all target sentences
as $\mathbf{t} \in \tau$. Now, we want our model to learn some probability
distribution $p_\theta(\mathbf{t}\mid\mathbf{s})$. What should
$p_\theta(\mathbf{t}\mid\mathbf{s})$ look like? We might define a loss function:
\begin{equation*}
  \begin{split}
    L_{\text{NLL}}(\theta) = - \sum_{i=1}^N \sum_{j=1}^J\sum_{k=1}^{\mid\nu\mid} \mathbbm{1}\{t_j = k\} \\
    \times \log p_\theta(t_j = k \mid \mathbf{s_i},t_{<j})
  \end{split}
\end{equation*}
Which is the usual auto-regressive negative-log likelihood. It's minimized when
$p_\theta(\mathbf{t}\mid\mathbf{s})$ assigns high probability to our observed
data. This is a good start.

\textbf{Word-level Knowledge Distillation}
In knowledge distillation \cite{hinton}, we are given a teacher model
$q(\mathbf{t}\mid\mathbf{s})$ that is already a decent approximation of our data
distribution $\mathcal{D}(\mathbf{t}\mid\mathbf{s})$. Usually the teacher is an
ensemble of models, or just a higher-capacity version our student. Since
$q(\mathbf{t}\mid\mathbf{s})$ is good, we want to craft a new loss function to
make $p_\theta(\mathbf{t}\mid\mathbf{s})$ look like
$q(\mathbf{t}\mid\mathbf{s})$. \newcite{Kim} give us two ways to do
this. If our teacher is auto-regressive, we can easily write:
\begin{equation*}
  \begin{split}
    L_{\text{WORD-KD}}(\theta) = - \sum_{i=1}^N \sum_{j=1}^J\sum_{k=1}^{\mid\nu\mid} q(t_j = k\mid\mathbf{s_i},t_{<j}) \\
    \times \log p_\theta(t_j = k \mid \mathbf{s_i},t_{<j})
  \end{split}
\end{equation*}
which minimizes the cross-entropy between $p_\theta$ and $q$ at each position
$j$ in the target sequence. This is called word-level knowledge distillation,
and now we can take some linear combination of $L_{\text{NLL}}$ and
$L_{\text{WORD-KD}}$ to learn from both our teacher and our data at the same
time.

\textbf{Sequence-level Knowledge Distillation}
It would be nice, however, to learn not only the local word distributions of
$q$, but also the sequence-level distributions. We want something like this:
\begin{equation*}
    L_{\text{SEQ-KD}}(\theta) = - \sum_{i=1}^N\sum_{\mathbf{t} \in \tau}
    q(\mathbf{t}\mid\mathbf{s_i}) \log p_\theta(\mathbf{t} \mid \mathbf{s_i})
\end{equation*}

which is a function of all of $q$, not just the values of $q$ conditioned on the
target prefixes we observe in training. Unfortunately, this exponential
summation is intractable.  \newcite{Kim} approximate $L_{\text{SEQ-KD}}$ with the mode of $q$, and approximate the mode with beam search. We encourage detail-oriented readers to refer
to the original paper for further explanation.

Finally, this gives us the sequence-level knowledge distillation procedure: (1)
train a big teacher model, (2) translate the source sentences with beam search,
(3) train a student model on the source sentence and teacher translation pairs.
Using this, \cite{Kim} show an improvement of up to 4 BLEU on WMT
English-German.\footnote{
There's also something called ``sequence-level interpolation'', which involves
taking a sample from the teacher beam that's close to the gold translation. For
simplicity, we'll ignore that here and perhaps return to it in future work.}

\section{Experiment Setup}

Before running our experiments, we face the non-trivial problem of selecting the appropriate hyper
parameters for our teacher and student models. How big is big enough? And at
what size does performance degrade? To our knowledge, the only decent way to
answer these questions on a given dataset is through grid search.

We start with an architecture known to produce reasonable results
on the TED German-English dataset\footnote{Downloaded from \url{http://www.cs.jhu.edu/~kevinduh/a/multitarget-tedtalks/}.} and grid search our way down.\footnote{Appendix Figure~\ref{fig:grid_search} shows the grid search results, plotting validation scores against number of parameters.}  We train each model for 100 epochs and measure the
tokenized BLEU validation score.

For our LARGE teacher architecture (9M parameters), we select a 1-layer encoder, 1-layer decoder Bi-LSTM architecture with 256 hidden and embed units. We randomly select a SMALL architecture (1.4M parameters) from those with less than 8M parameters\footnote{As a POC, we did the usual SLKD method on all student candidates and found an average improvement of 2.2 BLEU.} and use it as our student model.
  Interestingly, although the SMALL architecture has $\frac{1}{5}$th the number of parameters of the LARGE architecture, the only difference between the two is the BPE vocabulary size (10k vs. 500). 

Teachers are trained for 100 epochs. To ensure each student gets equal compute
time, we set the same max number of checkpoints (each 4000 updates) for models of the same
size. All models are trained using the Sockeye seq2seq framework \cite{sockeye}
and averaged over 3 trials.

\section{The Multi-Modal Hypothesis}


But why does SLKD perform well? \newcite{Kim} hypothesize that student models are too
low-capacity to fit the noisy training data and that by ``denoising'' the data,
capacity is freed up to model more important parts of translation space (which
are intuitively around the mode of the teacher distribution). While the mode of
the teacher distribution seems to have special properties, we challenge
the capacity assumption below.

A similar interpretation is that SLKD reduces
multi-modality in the data \cite{non-autoregressive}. The teacher simplifies the data
linguistically, replacing rare translations of phrases with more consistent
ones. They also assume that the model is too
low-capacity to deal with more complicated translations, and that by removing
``problematic'' examples, the model will only learn things that generalize well \cite{analyzing}.

Another explanation for why knowledge distillation works in general is
``dark knowledge'', a term coined by Hinton. However, this theory addresses the
case when the full teacher distribution is available to the student, which is
not true for SLKD. Furthermore, \cite{Kim} showed that word-level and sequence-level
distillation are roughly orthogonal and probably improve student models in
different ways.

Here, we examine the multi-modal hypothesis more closely. By constructing experiments with mixed multi-modal data, we
find that it fails to fully explain our empirical results.

\section{Testing the Multi-Modal Hypothesis}

We aim to answer the following question: does training on ``simplified'' SLKD
data free up capacity in our SMALL model, so that the model can focus on fitting
important parts of translation space? 

We train a LARGE model on the original training bitext (``base'' data) as an SLKD teacher and use it
to re-translate the TED source data, producing a ``kd'' dataset. 
We then compare training SMALL students on three kinds of data: base, kd, and the concatenation of the two (base+kd). 
Assuming the multi-modal hypothesis is true, we would expect that base+kd, being more varied than the original base dataset, would pose more challenges for the student.

\begin{table}[h]
\begin{center}
\begin{tabular}{c|ccc|c}
    Dataset & Trial 1 & Trial 2 & Trial 3 & Avg\\
    \hline
    base & 25.99 & 25.84 & 25.72 & 25.85 \\
    kd & 26.94 & 27.14 & 27.12 & 27.07\\
    base+kd & 26.96 & 27.49 & 27.39 & 27.28
\end{tabular}
\caption{Test BLEU of student models trained on different datasets for 30 checkpoints. Single trials with different student architectures are also plotted in Appendix Fig \ref{fig:concat_grid_search}.}
\end{center}
\end{table}

\begin{table}[t]
\begin{center}
\begin{tabular}{c|cc}
    & Avg \# Tok & Avg \# Vocab \\
    \hline
    base & 2.94M & 51k \\
    kd & 3M & 44k \\
    base+kd & 6.14M & 70k
\end{tabular}
\caption{\label{tab:stat} Average number of tokens and vocabulary size for baseline, kd, and base+kd target datasets.}
\end{center}
\end{table}

From our experiments, we see kd datasets indeed have smaller vocabularies (Table \ref{tab:stat}), but SMALL models have no problem fitting both ``noisy'' and ``denoised''
data concatenated together and can do so with equal amounts of compute time. Our difficulties with this TED data
lie not with the capacity of our students to fit
the data, but with our ability to find generalizable solutions.

\section{Data Augmentation Hypothesis}
We think that the multi-modal hypothesis is not the full explanation of SLKD performance; to explain the base+kd results, we additionally propose a data augmentation hypothesis: SLKD data adds something to datasets that makes it easier to learn generalizable solutions. We attempt to test and possibly improve this hypothesis with the following two additional data augmentation strategies and two different training conditions:

\textbf{Back-Translation}
We train a second LARGE teacher in the opposite direction
(English-German).\footnote{We call this teacher a ``backtranslation teacher''
  for brevity, although backtranslation typically refers to supplementing
  parallel MT data with mono-lingual data \cite{backtranslation}.} We translate
the TED target sentences with the backtranslation teacher to produce a third
``BT'' dataset, which we use to augment the KD and baseline datasets.

\textbf{Best 2}
We also experiment with using more of the teacher's beam. We produce two
versions of the data: one with the best translation from beam search and the
other with the second best. We combine these to form the best-2 dataset, which
we use to augment the baseline dataset.

\textbf{Longer Training}
We test if gains from SLKD can be recovered if baselines are given enough compute time. We train for 100 checkpoints, rather than 30, and find that our results remain unchanged.

\textbf{Less Dropout}
Finally, we test if adding SLKD data can replace the usual
regularization techniques. We experiment with turning off dropout on students,
which lets them overfit the training data.

\begin{table*}[ht]
  \begin{center}
    \begin{tabular}{|c|c|c|c|c|c|c|}
      \hline & \multicolumn{4}{|c|}{SMALL Students} & \multicolumn{2}{|c|}{LARGE Students}\\
      \hline
             & \multicolumn{2}{|c|}{w/ Dropout} & \multicolumn{2}{|c|}{No Dropout} & \multicolumn{2}{|c|}{}\\
      \hline Dataset & BLEU &  $\text{PPL}_{\text{Train}}$ & BLEU & $\text{PPL}_{\text{Train}}$ & BLEU & $\text{PPL}_{\text{Train}}$\\
      \hline
baseline & 26.79 & 4.86 & 25.37 & 4.24 & 31.75 & 4.99\\
kd & 27.70 & 2.17 & 26.45 & 2.09 & 30.38 & 1.93\\
base+kd & 27.74 & 3.53 & 27.84 & 3.02 & 32.52 & 3.33\\
base+kd+bt & 27.87 & 3.41 & \textbf{28.38} & 2.93 & \textbf{32.99} & 3.29\\
base+best-2 & \textbf{27.92} & 3.12 & 28.03 & 2.64 & 32.59 & 2.73\\
      \hline
    \end{tabular}
     
  \end{center}
  \caption{\label{aug-data-table} The tokenized test BLEU scores (Beam=5)\footnotemark\ and BPE train
    perplexities for student models trained on concatenations of datasets. SMALL students are trained for 100 checkpoints, rather than the initial 30.}
\end{table*}
\footnotetext{Contrary to the results observed by \cite{Kim}, we see significant gains when increasing the beam search size from 1 to 5.}

\begin{figure}
  \includegraphics[width=\columnwidth]{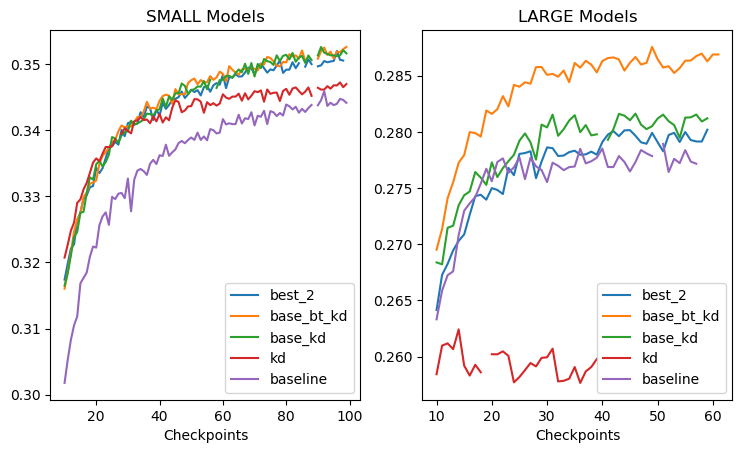}
  \caption{BPE validation curves of models trained on augmented datasets.}
  \label{val-curves}
\end{figure}

\section{Testing the Data Augmentation Hypothesis}
We show the results (BLEU and perplexity averaged over three trials) in Table ~\ref{aug-data-table}.
We see some
surprising results:

\begin{itemize}
\item{Further augmenting datasets does slightly improve performance in SMALL students.}
\item{Augmented datasets do not require more training time than baselines, but instead converge faster. (Figure ~\ref{val-curves})}
\item{Turning off regularization magnifies the gains from data augmentation in
    SMALL models, but hurts baseline and SLKD performance.}
\end{itemize}

It seems like SLKD can do the job of regularization, pulling SMALL models towards
simple, generalizable solutions. We see that when using this technique, more
data is better.

We further repeat the experiments using a LARGE student: this student has the same size and hyperparameters as the teacher (also known as BANs \cite{BAN})\footnote{Born Again Networks applied word-level knowledge distillation students of the same architecture, but to our knowledge, no one has attempted this with SLKD}, but is trained from scratch on the SLKD/augmented data. 
We observe that these augmented data also help LARGE students generalize better.

But how does SLKD regularize? In general there are two different ways of doing
regularization. The usual way is restricting the complexity of the model (via
dropout, L2, etc.). However, consider the case in which the true generating
function $G(x)$ of the data (or at least an approximation, like our teacher model) is known. An alternative way of regularizing would be to generate many more data
points along the manifold $G$ defines. Then any naive, overfitting model would
be naturally pulled towards the ``true'' solution.

Furthermore, we think this alternative way of regularizing is more appropriate for
model compression than dropout. While regularizing via dropout can help generalization, it
does so at the cost of model capacity.\footnote{Dropout has been characterized
  as training an ``ensemble of sub-networks'' \cite{understanding_dropout}.} Since our models are
already under-capacity, this hurts performance. Regularizing via SLKD, however,
helps the model generalize without restricting model capacity, which is why we
see such gains when turning off regularization.\footnote{But why
  does regularization hurt models trained on just SLKD? There might not be
  enough data to fully outline the manifold of $G(x)$. We think if we used the
  teacher on more data, this result would change.}

\section{Limitations and Future Work}

We investigated different hypotheses for why SLKD works and found that the conventional multi-modal hypothesis does not explain all the results. We proposed a complementary data augmentation hypothesis and showed that SLKD may work as an implicit regularizer.

We're continuing to experiment with different datasets, language pairs, and  model architectures (read: Transformers), where these results might change. We are interested in determining if the multi-modal hypothesis might still explain the gains from SLKD in any of these cases, or whether the regulatory effect of SLKD is the main contributor of increased performance in all cases.


\bibliography{emnlp-ijcnlp-2019.bib}
\bibliographystyle{acl_natbib}

\clearpage
\appendix
\section{Grid Search}
\begin{figure}[h!]
  \includegraphics[width=\columnwidth]{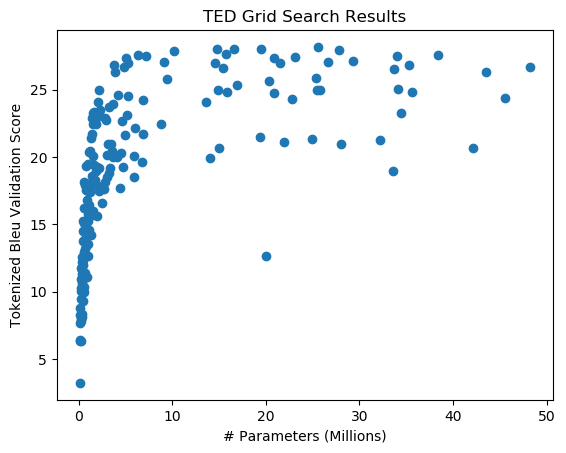}
  \caption{Grid search results on TED de-en over RNN hyperparameters. Note the
    performance drop around 5M parameters. This is analogous to some work done
    in network pruning, where models fail catastophically after passing some
    pruning threshold \cite{frankle2018the}.}
  \label{fig:grid_search}
\end{figure}

\begin{figure}[h!]
  \includegraphics[width=\columnwidth]{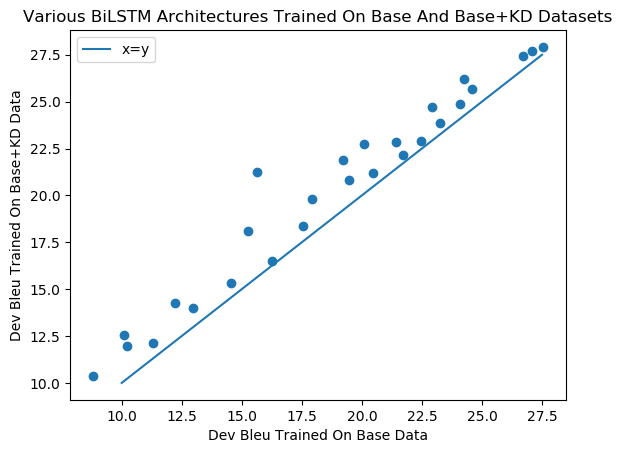}
  \caption{We train some 1-layer encoder 1-layer decoder Bi-LSTMs on both the base dataset and the concatenated base+kd dataset. Each point represents an architecture with different hyperparameters. All points are single trials. In all architectures, the performance of models trained on base+kd equals or exceeds the performance when trained on just base. BPE=[10000,750,500], num-hidden,num-embed=[64,128,256] }
  \label{fig:concat_grid_search}
\end{figure}

\subsection{Grid Search Parameters}
As supplied to sockeye.train:\\
``--num-words'' (BPE) = [30000, 20000, 10000, 7500, 5000, 1000, 750, 500, 250]\\
``--num-embed'' = [512, 256, 128, 64]\\
``--num-layers'' = [2, 1] (This how many layers in both the encoder and decoder \textbf{each}.)\\
``--rnn-cell-type'' = [lstm, gru]\\
``--rnn-num-hidden'' = [512, 256, 128, 64]\\

Note: some entries are missing in this grid. For example, for BPE less than
10000, we neglect training with embed size of 512 and num-hidden size of 512.
This was somewhat arbitrary, since the grid search was done in two separate
batches (none of the first batch showed a definite correlation between size and performance).

\subsection{Selected LARGE Architecture}
The model with the best BLEU validation score:\\
``--num-words'' (BPE) = 10000\\
``--num-embed'' = 256\\
``--num-layers'' = 1\\
``--rnn-cell-type'' = lstm\\
``--rnn-num-hidden'' = 256\\

\subsection{Selected SMALL Architecture}
Randomly selected from models with fewer than 8M parameters:\\
``--num-words'' (BPE) = 500\\
``--num-embed'' = 256\\
``--num-layers'' = 1\\
``--rnn-cell-type'' = lstm\\
``--rnn-num-hidden'' = 256\\

\subsection{Other Hyperparameters}

``--max-seq-len'' = ``100:100''\\
``--word-min-count'' = ``1:1''\\
``--checkpoint-frequency'' = ``4000''\\
``--batch-size'' = ``4096''\\
``--keep-last-params'' = ``3''\\
``--disable-device-locking'',\\
``--decode-and-evaluate'' = ``-1''\\
``--decode-and-evaluate-use-cpu''\\
``--initial-learning-rate'' = ``0.0003''\\
``--label-smoothing'' = ``0.1''\\
``--batch-type'' = ``word''\\
``--optimizer'' = ``adam''\\
``--gradient-clipping-threshold'' = ``1.0''\\
``--gradient-clipping-type'' = ``abs''\\
``--learning-rate-reduce-factor'' = ``0.7''\\
``--learning-rate-reduce-num-not-improved'' = ``8''\\
``--learning-rate-scheduler-type'' = \\``plateau-reduce''\\
``--learning-rate-decay-optimizer-states-reset'' = ``best''\\
``--learning-rate-decay-param-reset''\\
``--loss'' = ``cross-entropy''\\
``--embed-dropout'' = ``.0:.0''\\
``--encoder'' = ``rnn''\\
``--decoder'' = ``rnn''\\
``--rnn-attention-type'' = ``dot''\\
``--rnn-dropout-inputs'' = ``.1:.1''\\
``--rnn-dropout-states'' = ``.1:.1''\\

\subsection{Less Regularization}
When turning off dropout, we set\\
``--rnn-dropout-inputs'' = ``0:0''\\
``--rnn-dropout-states'' = ``0:0''\\

\end{document}